\documentclass[10pt,a5paper,twoside]{article}
\usepackage{coling2012}
\usepackage{subfig}
\usepackage{url}
\usepackage{slashbox}

\title{SpeedRead: A Fast Named Entity Recognition Pipeline}

\author{$Rami~Al-Rfou'~~~Steven~Skiena$ \\
  Department of Computer Science \\
  Stony Brook University \\
  NY 11794, USA \\
  \texttt{ \{ralrfou, skiena\}@cs.stonybrook.edu}}

\begin{document}
\maketitle
\abstractEn{
Online content analysis employs algorithmic methods to identify entities in
unstructured text. Both machine learning and knowledge-base approaches lie at
the foundation of contemporary named entities extraction systems. However, the
progress in deploying these approaches on web-scale has been been hampered by the
computational cost of NLP over massive text corpora. We present SpeedRead (SR), a
named entity recognition pipeline that runs at least 10 times faster than Stanford NLP pipeline. This pipeline consists of a high
performance Penn Treebank-compliant tokenizer, close to state-of-art part-of-speech (POS) tagger and knowledge-based named entity recognizer.
}

% TODO
\keywordsEn{Tokenization, Part Of Speech, Named Entity Recognition, NLP pipelines}

\newpage
\section{Introduction}

Information retrieval (IR) systems rely on text as a main source of data, which is processed using natural language processing (NLP) techniques to extract information and relations. Named entity recognition is essential in information and event-extraction tasks. Since NLP algorithms require computationally expensive operations, the NLP stages of an IR system become the bottleneck with regards to scalability \cite{DBLP:conf/acl/PaulsK11}.
Most of the relevant work, conducted by researchers, was limited to small corpora of news and blogs because of the limitation of the available algorithms in terms of speed. Most of the NLP pipelines use previously computed features that are generated by other NLP tasks, which adds computational cost to the overall NLP pipeline. For example, named entity recognition and parsing need POS tags; co-reference resolution requires named entities. In effect, we anticipate lower speed for future tasks.

A conservative estimate of a sample of the web news and articles can add up to terabytes of text. On such scale, speed makes a huge difference. For example, considering the task of annotating 10 TiBs of text with POS tags and named entities using a 20 CPU cores computer cluster would take at least 4 months using the fastest NLP pipeline available for researchers, our calculations show. Using our proposed NLP pipeline the time is reduced to a week.

\begin{figure*}[!h]
\begin{center}
\subfloat[POS taggers performance.]{\label{pospipes}\includegraphics[scale=0.35]{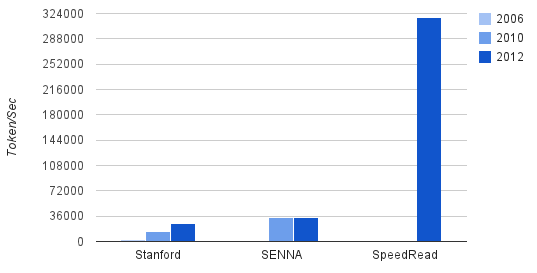}}
\subfloat[NER taggers performance.]{\label{nerpipes}\includegraphics[scale=0.35]{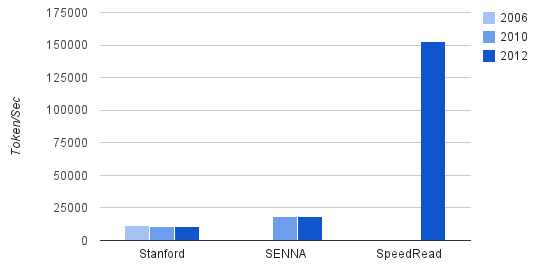}}
\caption{Performance of NLP pipelines through the years over POS and NER tagging. Stanford POS tagger uses L3W model, its speed in 2006 is slow to be apparent in the graph. Stanford tagger uses CONLL 4 classes model. SENNA pipeline was first released in 2008}
\label{pipes}
\end{center}
\end{figure*}

Several projects have tried to improve the speed by using code optimization. Figure \ref{pospipes} shows that Stanford POS tagger has improved throughout the years, increasing its speed by more than 10 times between 2006 and 2012. However, the current speed is twice slower than the SENNA POS tagger.

In this paper, we present a new NLP pipeline, SpeedRead, where we integrate global
knowledge extracted from large corpora with machine learning algorithms to
achieve high performance. Figures \ref{pospipes} and \ref{nerpipes} show that our pipeline is 10 times faster than Stanford pipeline in both tasks: POS tagging and NER tagging. Our design is built on two principles: (1) majority of the words have unique annotations and tagging them is an easy
task; (2) the features extracted for the frequent words should be cached for later use by the classifier. Both principles are simple and they show how to bridge the large gap in performance between current systems and what can be achieved.

\begin{table}[!h]
\centering
\begin{tabular}{ll}
\hline
\textbf{Phase} & \textbf{SpeedRead\,\,\,\,\,\,\,\,\,}\\
& \textbf{Relative Speed}\\
\hline
Tokenization & 11.8\\
POS & 11.1\\
NER & 13.9\\
TOK+POS+NER & 18.0\\
\end{tabular}
\caption{SpeedRead relative speed to Stanford pipeline.}
\label{}
\end{table}

Our work makes the following contributions:

\begin{compactitem}
\item \emph{Exposing the performance limitations of the current NLP systems:} We show that there is an algorithmic room for improving performance, rather than relying solely on optimizing the code.
\item \emph{High performance NLP pipeline that supports English tokenization, POS tagging and named entity recognition:} Novel design decisions that are not taken by most of the available tools to explore new area of the accuracy-performance space. SpeedRead is available under an open-source license. The code's organization is simple and it is written in Python for its readability benefits. This makes it easier for others to contribute and hack.
\item \emph{Techniques to reduce computation needed for sequence tagging tasks:} We distinguish between ambiguous and non-ambiguous words. We use the larger copora to calculate the frequent words and their frequent tags. We cache the extracted features of the most frequent words to avoid unnecessary calculations and boost performance.
\end{compactitem}

\begin{figure*}[t]
\begin{center}
\includegraphics[scale=0.65]{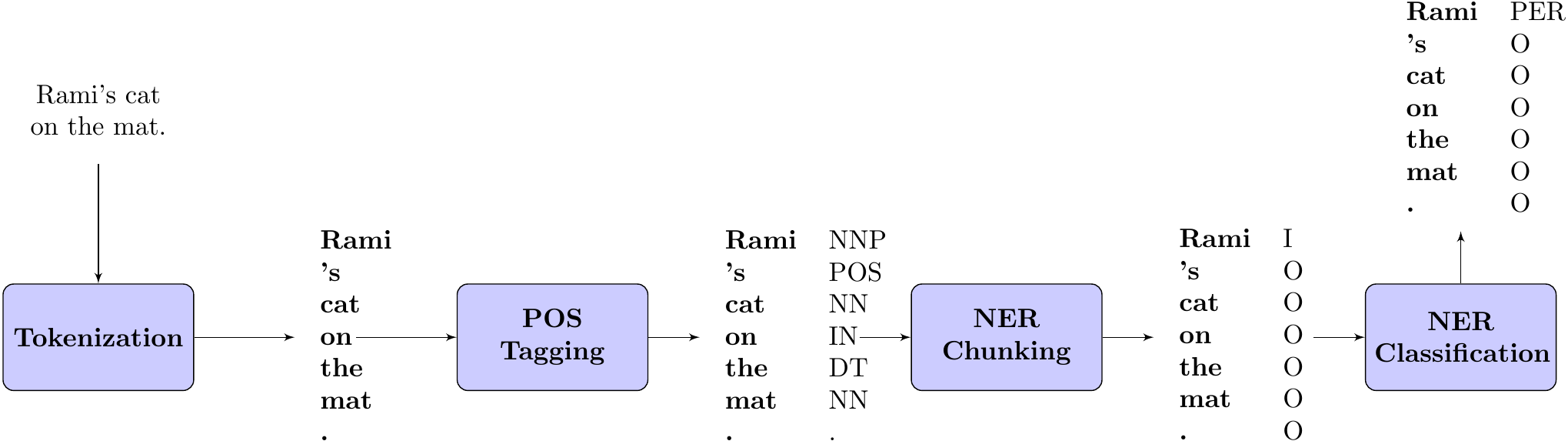}
\caption{SpeadRead named entity recognition pipeline. First, tokenization split the words into basic units to be processed in the later phases. POS tagging identifies to which speech categories words belong to. There are 45 part of speech category, we are mainly interested in nouns. Chunking identifies the borders of phrases that make up the named entities. In the above sentence, the named entity, Rami, is one word phrase. The last stage classifies each phrase to one of four categories; Person, Location, Organization or Miscellaneous.}
\label{arch}
\end{center}
\end{figure*}

Figure \ref{arch} shows the design of the SpeedRead pipeline. The first stage is tokenization followed by POS tagging that is used as an essential feature to decide the boundaries of the named entities' phrases. Once the phrases are detected, a classifier decides to which category these named entities belong to.

This paper is structured as follows. In Section \ref{related}, we discuss the current NLP pipelines, available to researchers.
Section \ref{tok} discusses SpeedRead tokenizer's architecture, speed and accuracy.
In Section \ref{POS}, we discuss the status of the current state-of-art POS taggers and describe SpeedRead new POS tagger. Section \ref{ner} describes the architecture SpeedRead's named entity recognition phase. Finally, in Section \ref{conc}, we discuss the status of the pipeline and the future improvements.

\subsection{Experimental Setup}
All the experiments presented in this paper were conducted on a single machine that has i7 intel 920 processor running on 2.67GHz, the operating system used is Ubuntu 11.10. The time of execution is the sum of \emph{\{sys, user\}} periods calculated by the Linux command \verb+time+. The speeds that are reported are calculated by averaging the execution time of five runs without considering any initialization times.

\section{Related Work}
\label{related}
There are many available natural language processing packages available for
researchers under open source licenses or non-commercial ones. However, this section is not meant to review the literature of named entity recognition research as this is already available in \cite{Nadeau}. We are
trying to discuss the most popular solutions and the ones we think are
interesting to present.

Stanford NLP pipeline \cite{Toutanova2000,Toutanova2003,klein2003conll,Finkel05,lee11conllst}
is one of the most popular and used NLP packages. The
pipeline is rich in features, flexible for tweaking and supports many
natural languages. Despite being written in Java, there are many
other programming language bindings that are maintained by the community. The pipeline
offers a tokenization, POS tagging, named entity recognition,
parsing and co-referencing resolution.
The pipeline requirements of memory and computation are non-trivial. To accommodate the various computational resources, the pipeline offers several models for each task that vary in speed, memory consumption and accuracy. In general, to
achieve good performance in terms of speed, the user has to increase the memory
available to the pipeline to 1-3 GiBs and choose the faster but less accurate models.

More recent efforts include SENNA pipeline. Even though it lacks a proper tokenizer,
it offers POS tagging, named entity recognition, chunking,
semantic role labeling\cite{Collobert:2008:UAN:1390156.1390177} and parsing \cite{Collobert_AISTATS_2011}. The pipeline has simple interface, high
speed and small memory footprint (less than 190MiB).

SENNA builds on the idea of deep learning of extracting useful features from unlabeled text. This unsupervised learning phase is done using auto-encoders and neural networks language models. It allows the pipeline to map words into another space of representation that has lower dimensionality. SENNA maps every word available in its 130 thousand word dictionary to a vector of 50 floating numbers. These vectors are then merged into a sentence structure using convolutional networks. The same architecture is then trained on different tasks using annotated text to generate different classifiers. The big advantage of taking this approach is the lesser amount of engineering that it requires to solve multiple problems.

NLTK \cite{BirdKleinLoper09} is a set of tools and interfaces to other NLP packages. Its simple APIs and
good documentation makes it a favorable option for students and researchers. Written
in Python, NLTK does not offer great speed or close to state-of-art
accuracy with its tools. On the other hand, it is well maintained and has great
community support.

WikipediaMiner \cite{Milne08aneffective} detects conceptual words and named entities; it also disambiguates the word senses. This approach can be modified to detect only the words that represent entities, then
using the disambiguated sense, it can decide which class the entity belongs to. Its
use of the Wikipedia interlinking information is a good example of the power
of using knowledge-based systems. Our basic investigation shows that the current
system needs large chunks of memory to load all the interlinking graph of Wikipedia and it would be hard to optimize for speed. TAGME \cite{Ferragina} is extending the work of WikipediaMiner to annotate short snippets of text. They are presenting a new disambiguation system that is faster and more accurate. Their system is much simpler and takes into account the sparseness of the senses and the possible lack of unambiguous senses in short texts.

Stanford and SENNA performed the best in terms of speed and quality in our early
investigation. Therefore, we will focus on both of them from now on as good representatives of a wide range of NLP packages.

\section{Tokenizer}
\label{tok}
The first task that an NLP pipeline has to deal with is tokenization and sentence segmentation \cite{tokfirst}. Tokenization target is to identify tokens in the text. Tokens are the basic units which need not to be processed in the subsequent stages.  Part of the complexity of tokenization comes from the fact that the definition of what a token is, depends on the application that is being developed. Punctuation brings another level of ambiguity; commas and periods can play different roles in the text. For example, we do not need to split a number like $1,000.54$ into more units whereas we need to split a comma-separated list of words. On the other hand, tokenization is important as it reduces the size of the vocabulary and improves the accuracy of the taggers by producing similar vocabulary to the one used for training.

As many NLP tasks' gold standards are dependent on Penn Treebank(PTB), a corpus of annotated text and parsed sentences taken from Wall Street Journal (WSJ), we opted for their tokenization scheme.

Searching for good tokenizers, we limited our options to the ones that support Unicode. We believe that Unicode support is essential to any applications that depends on the pipeline. Stanford tokenizer and Ucto \cite{UctoProject} projects offer almost Penn Treebank (PTB) compliant tokenizers plus other variations that are richer in terms of features.

Table \ref{tokspeed} shows that there is a substantial gap in performance between basic white space tokenizer (words are delimited by spaces or tabs and sentences are split by new line characters) and more sophisticated tokenizers as Stanford tokenizer and Ucto. We observed that the Stanford tokenizer is 50 times slower than the baseline (WhiteSpace tokenizer), which motivated us to look at the problem again.

\begin{table}[h!]
\begin{center}
\begin{tabular}{lrr}
\hline
	\textbf{Tokenizer} & \multicolumn{1}{l}{\textbf{Word/Second}} & \multicolumn{1}{l}{\textbf{Relative Speed}} \\
\hline
	Ucto & 185,500&0.8\\
	PTB Sed Script &214220&0.96\\
	Stanford & 222,176&1.0\\	
	SpeedRead & 2,626,183&\textbf{11.8}\\
	WhiteSpace & 11,130,048&50.0 \\

\end{tabular}
\caption{Speed of different tokenizers measured as word/second; Every tokenizer generates different number of tokens. For consistency, the original words count before tokenization used to calculate the speed. Words count is calculated using linux command \emph{wc}. Execution time includes both tokenization and sentence segmentation times with the exception that the original PTB Sed Script does not do sentence segmentation. Ucto's default configuration is used. Stanford tokenizer runs with strict PTB flag turned on.}
\label{tokspeed}
\end{center}
\end{table}

The Stanford tokenizer is implemented using JFlex, a Java alternative to Flex. The tokenizer matured over the years by adding more features and modes of operation which makes it harder for us to modify. Ucto uses C++ to compile a list of regular expressions that passes over the text multiple times.

SpeedRead, like the Stanford tokenizer, uses a lexical analyzer to construct the tokenizer. However, we use different generating engine than the (F)lex family. SpeedRead depends on Quex \cite{QuexProject}, a lexical analyzer generator, to generate our tokenizer. Quex makes different trade-off decisions than the usual lex tools when it comes to the tokenizer's generation time. Quex spends more time optimizing its internal NFA to produce a faster engine. While generating a tokenizer from a normal lex file can take few minutes, Quex takes hours for the same task. However, Quex supports Unicode in multiple ways and has similar description language to lex, but is cleaner and more powerful. The extensive multiple mode support makes it easy to write the lexical rules in understandable and organized way. All of that results in a fast C implementation of a Penn Treebank compliant tokenizer as Table \ref{tokspeed} shows.

As a design decision, we did not support some features which we believe will not affect the accuracy of the tokenizer. Table \ref{tokerr} shows the features which are not implemented. While some of the features are easy to add as supporting contractions, others, involving abbreviations especially \emph{U.S.}, prove to be complex \cite{Gillick:2009:SBD:1620853.1620920}.
\begin{table}[h!]
\begin{center}
\begin{tabular}{llll}
\hline
	\textbf{Feature} & \textbf{Text} & \textbf{PTB} & \textbf{SpeedRead}\\
\hline
	Reordering & Japan. ...& Japan ... . & Japan . ...\\
	Punctuation& U.S." & U.S. . " & U.S. " \\
	addition&&& \\
	Contractions& gimme & gim me & gimme \\

\end{tabular}
\caption{Some features that are not implemented in SpeedRead Tokenizer. Contractions that involves apstrophes are implemented in SpeedRead. For instance, \emph{can't} will be tokenized to \emph{ca n't}.}
\label{tokerr}
\end{center}
\end{table}

Table \ref{tokacc} shows that the accuracy of our tokenizer is Penn Treebank compliant, despite the missing features. Moreover, running SpeedRead and Stanford tokenizers over Reuters RCV1 corpus results in approximately 214, 215 million tokens consecutively.

\begin{table}[h!]
\begin{center}
\begin{tabular}{lr}
\hline
	\textbf{Tokenizer} & \textbf{Accuracy}\\
\hline
	PTB Sed Script & 100.0\%\\ 
	Stanford tokenizer & 99.7\%\\ 
	SpeedRead & \textbf{99.0}\%\\  
	White Space & 0.0\% \\

\end{tabular}
\end{center}
\caption{Accuracy of the tokenizers over the first 1000 sentence in the Penn Treebank. The gold standard was created by getting the tokenized text from the parse trees and manually segment the original text into sentences according to the parse trees. Errors in differentiating between starting and ending quotations are not considered. Not supporting MXPOST convention, replacing brackets with special tokens, is not considered necessary.}
\label{tokacc}
\end{table}

\subsection{Sentence Segmentation}

While PTB offers a set of rules for tokenization, their tokenizer assumes that the sentences are already segmented, which is done manually. SpeedRead's sentence segmentation uses the same rules that Stanford tokenizer uses. For instance, a period is an end of a sentence unless it is part of an acronym or abbreviation. The list of rules to detect those acronyms and abbreviations are taken from the Stanford tokenizer. Any quotations or brackets, that follow the end of the sentence, will be part of that sentence. Running SpeedRead's sentence segmentation on Reuters RCV1 generated 7.8 million sentences, while Stanford tokenizer generated 8.2 million sentences.

\section{Part of Speech Tagger (POS)}
\label{POS}
Earlier work to solve the POS tagging problem relied on lexical and local features using maximum entropy models \cite{Toutanova2000}. Later, more advanced models took advantage of the context words and their predicted tags \cite{Toutanova2003} to achieve higher accuracy. As POS tagging is a sequence tagging problem, modeling the sequence into a Maximum Entropy Markov Model (MEMM) or Conditional Random Fields (CRF) model (to infer the probability of the tags' sequences) seems to be the preferred option. The probability of each tag is computed using log-linear model with features that include large enough context words and their already-computed tags. This transforms every instance of the problem into a large vector of features that is expensive to compute. Then the sequence of vectors are fed to graphical model to compute the probability of each class, using the inference rules. The size of features' vector and the inference computation are the same regardless of the complexity of the problem.

Although the previous algorithms are sufficient to achieve satisfying accuracy, their computation requirements are overkill for most of the cases faced by the algorithm. For example, \emph{the} has a unique POS tag that never changes depending on its position in the sentence. Moreover, \emph{more} and \emph{that} are frequent enough in the English text that there is a need to cache their extracted features.

\subsection{Algorithm}
SpeedRead takes advantage of the previous observations and tries to distinguish between ambiguous and certain words. To understand such influences, we ran a Stanford POS tagger (left 3 words Model (L3W); trained on Wall Street Journal(WSJ), Sections 1-18) over a 1 GiB of news text to calculate the following dictionaries:

\begin{compactitem}
\item The most frequent POS tag of each token (Uni).
\item The most frequent POS tag of each token, given the previous POS tag (Bi).
\item The most frequent POS tag of each token, given the previous and next POS tags (Tri).
\end{compactitem}

Using the above dictionaries to calculate the POS tag of a word, leads to various precision/recall scores. \cite{lee11conllst} shows that using sieves is the solution to combine several rules/dictionaries. In a sieve algorithm, there is a set of rules that are cascaded after each other. The algorithm runs the rules from the highest in precision to the lowest. The first rule, matching the problem instance, returns its computed tag immediately. SpeedRead implements few sieves in the following order:

\begin{compactenum}
\item \textbf{Certain tokens}: Given a sentence, if the percentage frequency of the most frequent tag of a token is more than a threshold (in our work, 95\%) then return that tag.
\item \textbf{Left and Right tags} (Tri): For each token with unknown tag, return the most frequent tag, given the left and right POS tags if they are known.
\item \textbf{Left tags} (Bi): For each token with unknown tag, return the most frequent tag, given the left POS tag if it is known from the previous stages.
\item \textbf{Token tag} (Uni) : For each token with unknown tag up to this stage, return the most frequent tag.
\item \textbf{Backoff tag}: If the token is unknown, use regular expression tagger to deduce the tag; the regular expression tagger relies heavily on matching suffixes.
\end{compactenum}

\subsection{Results}
Table \ref{POSacc} shows the performance of different algorithms running on different sections of PTB. Stanford and SENNA models use sections 1-18, 19-21, 22-24 for training, development and testing datasets, respectively. Despite the simplicity of our algorithm, it achieves relatively high accuracy on the various datasets available.

Applying more context-aware rules, SpeedRead with sieves 1-5 (SR[Tri/Bi/Uni]) implemented, shows improvement in accuracy by around 2.85\% compared to just using unigrams, SpeedRead with sieves 1,4-5 (SR[Uni]). To be sure that our algorithm is robust enough and not overfitting the dataset, we calculated the dictionaries again by running SENNA POS tagger\cite{collobert:2011b} over Reuters RCV1 corpus and the results were similar.

\begin{table}[h!]
\begin{center}
\begin{tabular}{|l|lll|}
\hline
	\backslashbox[0pt][l]{POS Tagger}{Sections} & 19-21 & 22-24 & 1-24\\
\hline
	Stanford Bidirectional&  97.27& 97.32& 98.16\\

	Stanford L3W& 96.97 & 96.89 & 97.90\\

	SENNA & 97.81 & 96.99 & 97.68\\

	SR[Tri/Bi/Uni] & 96.73 & \textbf{96.39} & 96.66\\

	SR[Bi/Uni] & 96.06 & 95.82 & 96.03\\

	SR[Uni] & 93.73 & 93.56 & 93.70\\	
\hline	
\end{tabular}
\caption{Accuracy of different taggers on different sections of Penn Treebank. The first column corresponds to the development set and the second to the testing set.}
\label{POSacc}
\end{center}
\end{table}

Tables \ref{POSacc} and \ref{POSspeed} show the tradeoff between accuracy and speed. Stanford pipeline offers two models with different speeds and accuracies. Since Left 3 Words model (L3W) is the preferred tagger to use in practice, we chose it to be our reference in terms of speed. L3W model runs 18 times faster than the state-of-art Bidirectional model and is only 0.4\% less accurate. SpeedRead pushes the speed by another factor of 11 with only 0.5\% drop in accuracy. Since the speed of some algorithms vary with the memory used, every algorithm was given enough memory that adding more memory will not affect its speed. The memory footprint is reported in the fourth column of Table \ref{POSspeed}.

\begin{table}[h!]
\begin{center}
\begin{tabular}{lrrl}
\hline
	\textbf{POS Tagger} & \multicolumn{1}{l}{\textbf{Speed}}&\textbf{Relative}&\textbf{Memory}\\
			    & Token/Sec&\multicolumn{1}{l}{\textbf{Speed}}&in MiB\\
\hline
	Stanford Bi& 1389&0.04& 900\\

	Stanford L3W& 28,646&1.00&450\\

	SENNA & 34,385&1.20&150\\

	SR [Tri/Bi/Uni] & 318,368&\textbf{11.11}&600\\

	SR [Bi/Uni] & 397,501&13.87&250\\	

	SR [Uni] & 564,977&19.72&120\\

\end{tabular}
\caption{Speed of different POS taggers. The first two taggers are Stanford taggers. The first tagger runs the Bidirectional(Bi) model and the second runs the Left 3 Words (L3w) model. SpeedRead has three variations}
\label{POSspeed}
\end{center}
\end{table}

\subsection{Error Analysis}
\begin{figure}
\begin{center}
\includegraphics[scale=0.45]{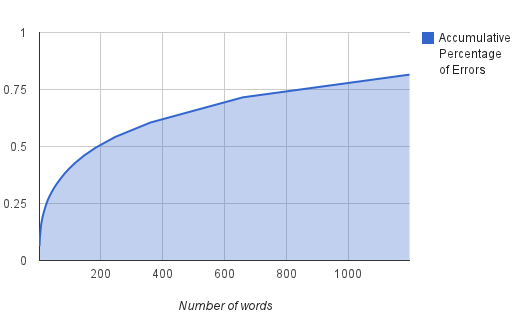}
\caption{Accumulative percentage of errors made by the most frequent mistagged words. The total number of words is around 2000, the graph lists only the most frequent 1000.}
\label{posdist}
\end{center}
\end{figure}

The most common errors are functional words, such as \emph{that, more, ..} which have multiple roles in speech. This confirms some of the conclusions reported by \cite{DBLP:conf/cicling/Manning11}. Figure \ref{posdist} shows that less than 10\% of mistagged words are responsible for slightly more than 50\% of the errors. Regarding unknown words, the only part of the tagger that generalizes over unseen tokens is the regular expression tagger. Regular expressions are not extensive enough to achieve high accuracy. Therefore, we are planning to implement another backoff phase for the frequent unseen words where we accumulate the sentences, containing these words, after sufficient amount of text is processed and then run Stanford/SENNA tagger over those sentences to calculate the most common tag.

Table \ref{cfm_pos} shows the confusion matrix of the most ambiguous tags; the less ambiguous tags are clustered into one category, \verb+O+. One of the biggest sources of confusion in tagging is between adjectives (JJ) and nouns (NN). Proper nouns are the second source of errors as most of the capitalized words will be mistakenly tagged as proper nouns while they are either adjectives or nouns. Such errors are the result of the weak logic implemented in the backoff tagger in SpeedRead, where regular expressions are applied in sequence returning the first match. Other types of errors are adverbs (RB) and propositions (IN). These errors are mainly because of the ambiguity of the functional words. Functional words need deeper understanding of discourse, semantic and syntactic nature of the text. Taking into consideration the contexts around the words improves the accuracy of tagging. However, trigrams are still small to be considered sufficient context for resolving all the ambiguities.

\begin{table*}
\begin{tabular}{l|rrrrrrrrrrr}
\backslashbox[0pt][l]{Ref}{Test}& DT& IN& JJ& NN& NNP& NNPS& NNS& RB& VBD& VBG & O\\
\hline
DT &11094&   62& 3& 7&  3&  0&  0&     1&     0&     0& 13 \\
IN & 15&13329& 9&     1&     0&     0&     0&    88&     0&     0& 50 \\
JJ & 1& 11& 7461&  \textbf{257}&   130&     2&    10&    65&    38&    81&   159\\
NN & 1& 5&   \textbf{288}&17196&  111&     0&    18&    11&     2&   109&    93 \\
NNP &     8& 13& 118& 109& 12585&  264&    31&     8&     0&     2& 39 \\
NNPS &  0&     0&     0&  0&    70&  81&   16&     0& 0& 0&     0 \\
NNS &    0&     0&     1& 23&    20&    42& 7922&    0& 0& 0&  53 \\
RB &   17&   \textbf{281}&   103&    23&     8&     0&     0& 3892&    0&     1&    80 \\
VBD &     0&    0&     8& 5& 4&     0&     0&     0& 4311&    1&   232 \\
VBG &     0&     0&    25&104&5& 0&     0&    0&     0& 1799&    0 \\
O &   26&   163&   154&   172&  47 & 4&   107&    67&   174&  2&45707\\
\end{tabular}\
\caption{Confusion Matrix of the POS tags assigned by SpeedRead over the words of sections 22-24 of PTB. O represents all the other not mentioned tags.}
\label{cfm_pos}
\end{table*}

\section{Named Entity Recognition (NER)}
\label{ner}

Named entity recognition is essential to understand and extract information from text. Many efforts and several shared tasks, aiming to improve named entity recognition and classification, had been made; CONLL 2000/2003 \cite{tjongkimsang2003conll} are some of the shared tasks that addressed the named entity recognition task. We use CONLL 2003's definition of named entity recognition and
classification task.
CONLL 2003 defines the chunk borders of an entity by using IOB tags, where
\verb+I-TYPE+ means that the word is inside an entity, \verb+B-TYPE+ means a beginning of a new entity if the
previous token is part of an entity of the same type and \verb+O+ for anything that is not part
of an entity. For classification, the task defines four different types:
Person(PER), Organization(ORG), Location(LOC) and Miscellaneous(MISC) (See Figure \ref{NER_task}).

We split the task into two phases. The first is to
detect the borders of the entity phrase. After the entity chunk is detected, the second phase
will classify each entity phrase to either a Person, Location, Organization or Miscellaneous.

\begin{figure}[!h]
\centering
\textbf{Columbia}/ORG \emph{is an} \textbf{American}/Misc \emph{university located in} \textbf{New}/LOC \textbf{York}/LOC.
\caption{Annotated text after NER.}
\label{NER_task}
\end{figure}

\subsection{Chunking}

We rely on the POS tags of the phrase words to detect the phrase that constitute an entity. A word is considered to be a part of an entity: (1) if it is a demonym (our compiled list contains 320 nationalities),  (2) if one of the following conjunction words \{\verb+&+, \verb+de+, \verb+of+\} appearing in middle of an entity phrase or, (3) if its POS tag is \verb+NNP(S)+ except if it belongs to one of these sets:
\begin{compactitem}
\item Week days and months and their abbreviations.
\item Sports (our compiled list contains 182 names).
\item Job and profession titles (our compiled list contains 314 title).
\item Single Capital letters.
\end{compactitem}
These sets are compiled using freebase.

CONLL dataset shows a strong correlation between POS tags \verb+NNP(S)+ and the words that are part of entities' phrases; 86\% of the words that appear in entities' phrases have \verb+NNP(S)+ POS tags. The remaining words are distributed among different POS tags; 6.3\% are demonyms. Adding the demonyms and proper nouns guarantee 92.3\% coverage of the entities' words that appear in the dataset.

Using POS tags as main criteria to detect the entity phrases is expected, given the importance of the POS tags for the NER task. 14 out of 16 submitted paper to CONLL 2003 used POS tags as part of their feature set.

The behavior of the chunking algorithm is greedy as it tries to concatenate as many consecutive words as possible into one entity phrase. A technical issue appears in detecting the borders of phrases when multiple entities appear after each other without non-entity separator. This situation can be divided into two cases. Firstly, if the two consecutive entities are of the same type. In this case, the chunking tag should be \verb+B-TYPE+. Looking at the dataset, such tag appears less than 0.2\% out of all the entities' tags. For example, in the original Stanford MEMM implementation, the classifier \cite{klein2003conll} generates \verb+IOB+ chunking tags while in the later CRF models \cite{Finkel05} only \verb+IO+ chunking tags are generated. The second case is when the phrases are of different types. In the dataset, this case appears 248 times over 34834 entities. Since both cases are not frequent enough to harm the performance of the classifiers, SpeedRead does not recognize them.

\subsubsection{Results}
Table \ref{detacc} shows F1 score of the chunking phase using different taggers to generate the POS tags. This score is calculated over the chunking tags of the words. \verb+I+ and \verb+B+ tags are considered as one class while \verb+O+ is left as it is. It is clear from Table \ref{detacc} that using better POS taggers does not necessarily produce better results. The quality of SpeedRead POS tagging is sufficient for the chunking stage. SENNA and SpeedRead POS taggers work better for the detection phase because they are more aggressive, assigning the \verb+NNP+ tag to any capitalized word. On the other hand, Stanford tagger prefers to assign the tag of the lowered case shape of the word, if it is a common word.

\begin{table}[h!]
\begin{center}
\begin{tabular}{|l|lll|}
\hline
	\backslashbox[100pt][l]{Phase}{Dataset} & Train & Dev & Test\\
\hline
	SR+SR POS &  94.24& 94.49& \textbf{93.12}\\
	SR+Stanford POS L3W &  92.98& 93.37& 92.05\\
	SR+CONLL POS &  90.88& 90.82& 89.43\\
	SR+SENNA POS &  94.73& 95.07& \textbf{93.80}\\	
\hline
\end{tabular}
\caption{F1 scores of the chunking phase using different POS tags. F1 score is calculated over tokens and not entities.}
\label{detacc}
\end{center}
\end{table}

\subsubsection{Error Analysis}
Table \ref{deterr} shows the error cases that appears in the chunking phase. The most common class of errors in the chunking phase is titles, such as \emph{\{RESULTS, DIVISION, CONFERENCE, PTS, PCT\}}. These words seem to confuse the POS tagger. Another source of confusion for the POS tagger is the words \emph{\{Women, Men\}}; such words appear in the name of sports so they get assigned \verb+NNP+ tag. As expected, all numbers that are part of entities are not detected. Conjunction words are the second important class of errors. \cite{mazurconjunction} shows that conjunction words that appear in middle of entities' phrases are hard to detect and need special classification task. As most of \emph{of} occurrences  are part of entities and the converse is true for \emph{and}, we decided to include the former and exclude the later.

\begin{table}[h!]
\begin{center}
\begin{tabular}{lrl}
\hline
	Word & Percentage & Type of error \\
\hline
	Titles & 22.7\%&Detected\\
	Titles & 4.9\%&Missed\\	
	of  & 2.6\%&Detected\\
	{96, 95, 1000 ...}&2.6\%&Missed\\
	Men &  1.3\%&Detected\\
	Women &  1.3\%&Detected\\
	and&  1.1\%&Missed\\
	central&  1.1\%&Detected\\
\end{tabular}
\caption{Most frequent errors in the chunking stage.}
\label{deterr}
\end{center}
\end{table}

\subsection{Classification}
Classification is a harder problem than just detecting an entity. For example,
``West Bank" can belong to two classes, location and organization. Disambiguating the sense of an entity
depends on the context. For
instance, ``Mr. Green" indicates that ``Green" is a person, while ``around Green"
points to a location. To classify an entity, we used a logistic
regression classifier, sklearn \cite{sklearn}. The features we feed to the classifier are two factors per type: $\phi_{ij}(Type_i, phrase_j)$ and $\psi_{ij}(Type_i, context_j)$. Context consists of two words that precede and follow an entity phrase. To calculate these factors:

\begin{equation}
\phi_{ij}(Type_i, phrase) = \prod_{k}^{n} P(Type_i \vert w_k)
\end{equation}

\begin{equation}
\psi_{ij}(Type_i, context=\{w_{before},w_{after}\}) = 
P(Type_i \vert w_{before}) \times P(Type_i \vert w_{after})
\end{equation}

The conditional probabilities of the types, given a specific word, are calculated using the distribution of tags frequencies over words, retrieved from the annotated Reuters RCV1 corpus. SENNA NER tagger has been used to annotate the corpus.

Table \ref{neracc} indicates the importance of the classification phase. First row shows that, given chunked input, the classification phase is able to achieve close scores to the state-of- art classifiers. However, given the chunks generated by SpeedRead, the scores drop around 9.5\% in F1 scores.

\begin{table}[h!]
\begin{center}
\begin{tabular}{|l|l|l|l|}
\hline
	\backslashbox[80pt][l]{Phase}{Dataset} & Training & Dev & Test\\

\hline
	SR+Gold Chunks & 90.80& 91.98& 87.87\\

	SpeeRead & 82.05& 83.35& 78.28\\

	Stanford &99.28&92.98&89.03\\

	SENNA	&96.75&97.24&89.58 \\
\hline
\end{tabular}
\caption{F1 scores calculated using conlleval.pl script for NER taggers. The table shows that SpeedRead F1 score is 10\% below the sate-of-art achieved by SENNA.}
\label{neracc}
\end{center}
\end{table}

To analyze the scores of the classification phase further, Table \ref{cfm_ner} shows a confusion matrix over the tags generated by SpeedRead. The errors that involve \verb+O+ are signs of chunking errors; there are 1158 chunking errors which exceed the total number of classification errors, 849.

\begin{table}[h!]
\begin{center}
\begin{tabular} {|l|rrrrr|}
\hline
\backslashbox[0pt][l]{Ref}{Test} & LOC & MISC& ORG & PER & O\\
\hline
 LOC & 1737 &  34 &   95 &   36&   23  \\
MISC &    36 & 660 &   57 &   52 & 113  \\

 ORG &   \textbf{323}  & 73   &1954  & 37& 109 \\
 PER &    26  &  8   &   72 &2632 &  35 \\
  O &    66  & 248  & \textbf{412} &  152 &37445 \\
\hline
\end{tabular}
\end{center}
\caption{Confusion matrix of the SpeadRead NER tags over the CONLL test dataset tokens.}
\label{cfm_ner}
\end{table}

The chunking errors contain more false positives than false negatives. The chunking algorithm is aggressive in considering every \verb+NNP(S)+ as part of an entity. That would be fine if we had a perfect POS tagger. The reality that the POS tagger has hard time classifying uppercased words in titles and camel cased words that appear at the beginning of the sentence.

Once non-entity is considered part of an entity phrase, the classifier has higher chance of classifying it as an \emph{ORG} than any other tag. The names of the organizations contain a mix of locations and persons' names, forcing the classifier to consider any long or mix of words as an organization entity. That appears more clearly in the second most frequent category of errors. 323 words in organizations entities' names were classified as locations. This could be explained by the fact that many companies and banks name themselves after country names and their locations. For example, ``Bank of England" could be classified as a location because of the strong association between England and the tag location.

Table \ref{nerspeed} shows that Stanford pipeline has a high cost for the accuracy achieved by the classifier. SENNA achieves close accuracy with twice the speed and less memory usage. SpeedRead takes another approach by focusing on speed. We are able to speed up the pipeline to the factor of 13. SpeedRead's memory footprint is half the memory consumed by the Stanford pipeline. Even though SpeedRead's accuracy is not close to the state-of-art, it still achieves 18\% increase over the CONLL 2003 baseline. Moreover, adapting the pipeline to new domains could be easily done by integrating other knowledge base sources as freebase or Wikipedia. SENNA and SpeedRead are able to calculate POS tags at the end of the NER phase without extra computation while that is not true of Stanford pipeline standalone NER application. Using Stanford corenlp pipeline does not guarantee better execution time.

\begin{table} [h!]
\begin{center}
\begin{tabular}{lrrl}
\hline
	\textbf{NER Tagger} & \textbf{Token/Sec}&\textbf{Relative}&\textbf{Memory}\\
	&&\textbf{Speed}&MiB\\	
\hline
	Stanford & 11,612&1.00&1900\\
	SENNA& 18,579&2.13&150\\
	SpeedRead & 153,194&\textbf{13.9}&950\\

\end{tabular}
\caption{Speed of different NER taggers. SpeedRead is faster by 13.9 times using half the memory consumed by Stanford.}
\label{nerspeed}
\end{center}
\end{table}

\section*{Conclusion and Future Work}
\label{conc}
Our success in implementing a high performance tokenizer and POS tagger shows that it is possible to use simple algorithms and conditional probabilities, accumulated from a large corpora, to achieve good classification and chunking accuracies.

This could lead to a general technique of approximating any sequence tagging problem using sufficiently large dictionaries of conditional probabilities of contexts and inputs. This approximation has the advantage of speeding up the calculations and opens the horizon for new applications where scalability matters.

Expanding this approach to other languages depends on the availability of other high accurate taggers in these languages. We are looking to infer these conditional probabilities from a global knowledge base as freebase or the interlinking graph of Wikipedia.

SpeedRead is available under GPLv3 license and it is available to download from \url{www.textmap.org/speedread}. We anticipate that it will be useful to large spectrum of named entity recognition applications.

\bibliography{myrefs}{}

\begin{thebibliography}{}

\bibitem[\protect\astroncite{Bird et~al.}{2009}]{BirdKleinLoper09}
Bird, S., Klein, E., and Loper, E. (2009).
\newblock {\em {Natural Language Processing with Python}}.
\newblock O'Reilly Media.

\bibitem[\protect\astroncite{Collobert}{2011}]{Collobert_AISTATS_2011}
Collobert, R. (2011).
\newblock Deep learning for efficient discriminative parsing.
\newblock In {\em International Conference on Artificial Intelligence and
  Statistics}.

\bibitem[\protect\astroncite{Collobert and
  Weston}{2008}]{Collobert:2008:UAN:1390156.1390177}
Collobert, R. and Weston, J. (2008).
\newblock A unified architecture for natural language processing: deep neural
  networks with multitask learning.
\newblock In {\em Proceedings of the 25th international conference on Machine
  learning}, ICML '08, pages 160--167, New York, NY, USA. ACM.

\bibitem[\protect\astroncite{Collobert et~al.}{2011}]{collobert:2011b}
Collobert, R., Weston, J., Bottou, L., Karlen, M., Kavukcuoglu, K., and Kuksa,
  P. (2011).
\newblock Natural language processing (almost) from scratch.
\newblock {\em Journal of Machine Learning Research}, 12:2493--2537.

\bibitem[\protect\astroncite{Ferragina and Scaiella}{2010}]{Ferragina}
Ferragina, P. and Scaiella, U. (2010).
\newblock Tagme: on-the-fly annotation of short text fragments (by wikipedia
  entities).
\newblock In {\em Proceedings of the 19th ACM international conference on
  Information and knowledge management}, CIKM '10, pages 1625--1628, New York,
  NY, USA. ACM.

\bibitem[\protect\astroncite{Finkel et~al.}{2005}]{Finkel05}
Finkel, J.~R., Grenager, T., and Manning, C. (2005).
\newblock Incorporating non-local information into information extraction
  systems by gibbs sampling.
\newblock In {\em In ACL}, pages 363--370.

\bibitem[\protect\astroncite{Gillick}{2009}]{Gillick:2009:SBD:1620853.1620920}
Gillick, D. (2009).
\newblock Sentence boundary detection and the problem with the u.s.
\newblock In {\em Proceedings of Human Language Technologies: The 2009 Annual
  Conference of the North American Chapter of the Association for Computational
  Linguistics, Companion Volume: Short Papers}, NAACL-Short '09, pages
  241--244, Stroudsburg, PA, USA. Association for Computational Linguistics.

\bibitem[\protect\astroncite{Gompel}{2012}]{UctoProject}
Gompel, M.~V. (2012).
\newblock Ucto: Unicode tokenizer.
\newblock \url{http://ilk.uvt.nl/ucto}.

\bibitem[\protect\astroncite{Klein et~al.}{2003}]{klein2003conll}
Klein, D., Smarr, J., Nguyen, H., and Manning, C.~D. (2003).
\newblock Named entity recognition with character-level models.
\newblock In Daelemans, W. and Osborne, M., editors, {\em Proceedings of
  CoNLL-2003}, pages 180--183. Edmonton, Canada.

\bibitem[\protect\astroncite{Lee et~al.}{2011}]{lee11conllst}
Lee, H., Peirsman, Y., Chang, A., Chambers, N., Surdeanu, M., and Jurafsky, D.
  (2011).
\newblock Stanford's multi-pass sieve coreference resolution system at the
  conll-2011 shared task.
\newblock In {\em Proceedings of the CoNLL-2011 Shared Task}.

\bibitem[\protect\astroncite{Manning}{2011}]{DBLP:conf/cicling/Manning11}
Manning, C.~D. (2011).
\newblock Part-of-speech tagging from 97\% to 100\%: Is it time for some
  linguistics?
\newblock In Gelbukh, A.~F., editor, {\em CICLing (1)}, volume 6608 of {\em
  Lecture Notes in Computer Science}, pages 171--189. Springer.

\bibitem[\protect\astroncite{Milne and Witten}{2008}]{Milne08aneffective}
Milne, D. and Witten, I.~H. (2008).
\newblock An effective, low-cost measure of semantic relatedness obtained from
  wikipedia links.
\newblock In {\em In Proceedings of AAAI 2008}.

\bibitem[\protect\astroncite{Nadeau and Sekine}{2007}]{Nadeau}
Nadeau, D. and Sekine, S. (2007).
\newblock A survey of named entity recognition and classification.
\newblock {\em Lingvisticae Investigationes}, 30(1):3--26.

\bibitem[\protect\astroncite{Pauls and Klein}{2011}]{DBLP:conf/acl/PaulsK11}
Pauls, A. and Klein, D. (2011).
\newblock Faster and smaller n-gram language models.
\newblock In Lin, D., Matsumoto, Y., and Mihalcea, R., editors, {\em ACL},
  pages 258--267. The Association for Computer Linguistics.

\bibitem[\protect\astroncite{Pawel and Robert}{2007}]{mazurconjunction}
Pawel, M. and Robert, D. (2007).
\newblock Handling conjunctions in named entities.
\newblock {\em Lingvisticae Investigationes}, 30(1):49--68.

\bibitem[\protect\astroncite{Schafer}{2012}]{QuexProject}
Schafer, F.-R. (2012).
\newblock Quex - fast universal lexical analyzer generator.
\newblock \url{http://quex.sourceforge.net}.

\bibitem[\protect\astroncite{Scikit}{2011}]{sklearn}
Scikit, S.-l.~D. (2011).
\newblock Scikit-learn: Machine learning in python.
\newblock {\em Journal of Machine Learning Research}, 12:2825--2830.

\bibitem[\protect\astroncite{Tjong Kim~Sang and
  De~Meulder}{2003}]{tjongkimsang2003conll}
Tjong Kim~Sang, E.~F. and De~Meulder, F. (2003).
\newblock Introduction to the conll-2003 shared task: Language-independent
  named entity recognition.
\newblock In Daelemans, W. and Osborne, M., editors, {\em Proceedings of
  CoNLL-2003}, pages 142--147. Edmonton, Canada.

\bibitem[\protect\astroncite{Toutanova et~al.}{2003}]{Toutanova2003}
Toutanova, K., Klein, D., Manning, C.~D., and Singer, Y. (2003).
\newblock Feature-rich part-of-speech tagging with a cyclic dependency network.
\newblock In {\em Proceedings of the 2003 Conference of the North American
  Chapter of the Association for Computational Linguistics on Human Language
  Technology - Volume 1}, NAACL '03, pages 173--180, Stroudsburg, PA, USA.
  Association for Computational Linguistics.

\bibitem[\protect\astroncite{Toutanova and Manning}{2000}]{Toutanova2000}
Toutanova, K. and Manning, C.~D. (2000).
\newblock Enriching the knowledge sources used in a maximum entropy
  part-of-speech tagger.
\newblock In {\em Proceedings of the 2000 Joint SIGDAT conference on Empirical
  methods in natural language processing and very large corpora: held in
  conjunction with the 38th Annual Meeting of the Association for Computational
  Linguistics - Volume 13}, EMNLP '00, pages 63--70, Stroudsburg, PA, USA.
  Association for Computational Linguistics.

\bibitem[\protect\astroncite{Webster and Kit}{1992}]{tokfirst}
Webster, J.~J. and Kit, C. (1992).
\newblock Tokenization as the initial phase in nlp.
\newblock In {\em Proceedings of the 14th conference on Computational
  linguistics - Volume 4}, COLING '92, pages 1106--1110, Stroudsburg, PA, USA.
  Association for Computational Linguistics.

\end{thebibliography}
\bibliographystyle{apa}

\end{document}